\pdfoutput=1

\documentclass[11pt]{article}

\usepackage[]{EMNLP2022}

\usepackage{times}
\usepackage{latexsym}
\usepackage{setspace}
\usepackage{multicol}
\usepackage{multirow}
\usepackage{graphicx}
\usepackage[T1]{fontenc}

\usepackage[utf8]{inputenc}

\usepackage{microtype}

\usepackage{xcolor}

\usepackage{inconsolata}

\title{Guiding Generative Language Models for \\ Data Augmentation in Few-Shot Text Classification}

\author{Aleksandra Edwards$\dagger$ 
\qquad Asahi Ushio$\dagger$ 
\qquad Jose Camacho-Collados$\dagger$
\\ \bf{H\'el\`ene de Ribaupierre}$\dagger$
\qquad   \bf{Alun Preece}$\ddag$
\\
$\dagger$School of Computer Science and Informatics, Cardiff University, United Kingdom \\$\ddag$Crime and Security Research Institute, Cardiff University, United Kingdom\\
\normalsize \tt\{edwardsai,ushioa,camachocolladosj,deribaupierreh,preecead\}@cardiff.ac.uk
}

\begin{document}
\maketitle
\begin{abstract}
Data augmentation techniques are widely used for enhancing the performance of machine learning models by tackling class imbalance issues and data sparsity. State-of-the-art generative language models have been shown to provide significant gains across different NLP tasks. However, their applicability to data augmentation for text classification tasks in few-shot settings have not been fully explored, especially for specialised domains. In this paper, we leverage GPT-2~\cite{radford2019language} for generating artificial training instances in order to improve classification performance. Our aim is to analyse the impact the selection process of seed training examples has over the quality of GPT-generated samples and consequently the classifier performance. We propose a human-in-the-loop approach for selecting seed samples. Further, we compare the approach to other seed selection strategies that exploit the characteristics of specialised domains such as human-created class hierarchical structure and the presence of noun phrases. Our results show that fine-tuning GPT-2 in a handful of label instances leads to consistent classification improvements and outperform competitive baselines. The seed selection strategies developed in this work lead to significant improvements over random seed selection for specialised domains. We show that guiding text generation through domain expert selection can lead to further improvements, which opens up interesting research avenues for combining generative models and active learning. 
\end{abstract}

\section{Introduction}

Data sparsity and class imbalance are common problems in text classification tasks \cite{turker2019knowledge,zhang2015short,shams2014semi,kumar2020data}, especially when the text to be labelled is from a highly-specialised domain where only scarce domain experts can perform the labelling task \cite{turker2019knowledge,ali2019text,lu2021sentence}. Data Augmentation (DA) is a widely used method for tackling such issues ~\cite{anaby2020not,kumar2020data,papanikolaou2019data}.  However,  the well-established DA methods in domains such as computer vision and speech recognition~ \cite{anaby2020not,giridhara2019study,10.1145/3065386,cui2015data,ko2015audio,szegedy2015going}, relying on simple transformations of existing samples, cannot be easily transferred to textual data as they can lead to syntactic and semantic distortions to text~\cite{giridhara2019study,anaby2020not}.  

Recent advances in text generation models, such as GPT and subsequent releases \cite{radford2018improving}, have led to the development of new DA approaches which generate additional training data from original samples, rather than perform only local changes to the text. Related studies use text generation models for improving relation extraction \cite{papanikolaou2019data,kumar2020data}, tackle class imbalance problems for extreme multi-label classification tasks \cite{zhang2020data}, and augment domain-specific datasets in order to improve performance in various domain-specific classification tasks \cite{amin2020exploring}. Specifically, \newcite{kumar2020data} and \newcite{anaby2020not} explore different fine-tuning approaches for pre-trained models for data augmentation in order to preserve class-label information. Results showed the potential of generative models such as GPT-2~\cite{radford2019language} and BART \cite{lewis2019bart} to augment small collections of labelled data. Further, an important problem with text generation techniques is the possibility of generating noise which decreases the performance of classification models rather than improving it~\cite{yang2020g}. However, this problem is ignored in the aforementioned studies.

The most similar study to ours is that of \newcite{yang2020g} in the context of commonsense reasoning. They proposed an approach based on the use of influence functions and heuristics for selecting the most diverse and informative artificial samples from an already-generated artificial dataset. Instead, we focus on the previous step of selecting the most informative samples (or \textit{seeds}) from the original data. We show that a careful selection of class representative samples from the original data in the first place can already lead to improvements and has an important efficiency advantage, as it prevents an unnecessary waste of resources and time of generating unused generated documents, especially considering how resource expensive generative language models are \cite{strubell-etal-2019-energy,schwartz2019green}. Finally, there is no research on exploiting the use of experts knowledge for improving the performance of generative language models for specialised domains.

Therefore, our aim is to improve the quality of generated artificial instances used for text classification training by developing seed selection strategies to guide the generation process. Specifically, we propose three DA methods in order to improve few-shot text classification performance using GPT-2 --- 1) a human-in-the-loop method that involves a domain expert choosing class representative samples; 2) a method that leverages the expert-generated classification hierarchy of a dataset in order to improve the classification of the top hierarchy classes; 3) a method that selects the seeds with the maximum occurrence of nouns. We chose these seed selection strategies because they exploit characteristics associated with specialised domains such as high number of terms, annotation performed by experts, and hierarchical class structure (common for social science and medical domains which require thematic analysis).


Our contributions are summarised as follows.
\begin{itemize}
    \item We advocate an important but not-well-studied problem of exploring how the quality of generated data and consequently few-shot classification can be improved using text generation-based DA strategies. We perform analysis for more specialised domain requiring domain experts for annotation.
    \item We propose novel seed selection strategies and analyse their impact on the performance of text generation-based data augmentation methods for few-shot text classification --- We show that classification performance can be improved significantly for specialised domains with limited labelled data using seed selection strategies and label preservation techniques. The human-in-the-loop seed selection proved to be the most suitable method for improving the quality of the generated data for specialised domains.
    \item We analyse how different approaches of fine-tuning GPT-2 model affect the quality of generated data and consequently the classification performance.
    
\end{itemize}

\section{Methodology}\label{sec:methodology}
     
     We experiment with two fine-tuning techniques for GPT in order to identify optimal ways for adapting GPT-2 model for DA for classification.  Further, our analysis focus on few-shot classification because of the demand for approaches which can perform well for only a handful of training instances especially in specialised domains where experts are sparse and data access is limited. However, our methodology can be easily extended for classification problems with more labelled data and it can also be used to generate more artificial training data.
\subsection{Seed Selection Strategies}\label{seedselection}
   
    We implement four seed selection strategies, which we describe below. 
    \paragraph{Human-in-the-loop Seed Selection.}\label{expertselection}
    The highly specialised nature of some domains where the manual annotation of documents is performed by experts show that identifying class representative samples might require more implicit knowledge that is hard to be captured by statistical approaches. 
    Therefore, we conducted a study asking experts to select the class representative samples from the original training data. The chosen seeds are then used to generate additional training data. We explain the approach in Section~\ref{ssec:case}.
    \paragraph{Maximum Nouns-guided Seed Selection.}\label{nounsselection} Many specialised domains are rich of domain-specific terminology and thus we believe that noun-rich instances might be more indicative for the classes compared to the other training samples. Therefore, we use this strategy to select the seeds with the maximum occurrence of nouns. We identify  single word nouns and compound nouns within data using NLTK~\cite{bird-loper-2004-nltk}.
    \paragraph{Subclass-guided Seed Selection.}\label{subclassselection} In this strategy, we leverage the human-generated classification hierarchy of a dataset in order to improve the classification of the top classes. Specifically, we select a roughly balanced number of seeds from each subclass belonging to a given label. In this way, we diversify the vocabulary for each overall class by ensuring the equal participation of representative samples from even the most underrepresented subclasses. 
    \paragraph{Random Seed Selection.}\label{randomselection} For this strategy we simply select a fixed number of instances in a random manner. We use random selection to evaluate whether the rest of the seed selection strategies lead to improvements in classification.

\subsection{Text Generation}\label{tgstrategy}
We generate artificial data using the generative pre-trained model, GPT-2 \cite{radford2019language}. We use GPT-2 model as it gives a state-of-the-art performance for many text generation tasks and also have been designed with the objective to fit scenarios with few-shot and even zero-shot settings. We use two methods for fine-tuning the GPT-2 model --- we fine-tune the model on the entire dataset and we also fine-tune a specific GPT-2 model for each given class to ensure label-preservation for the generated sequences. Fine-tuning a separate GPT-2 model per label ensures that each model has been exposed to text associated with a single class. We also perform experiments using a pre-trained GPT-2 model. We compare three models in order to assess the need of fine-tuning and the use of additional methods for label-preservation when using TG-based DA for classification tasks. These models are then leveraged to generate new documents given a labeled instance. These analyses help identify whether fine-tuning a separate model per label is a suitable method for ensuring label-preservation of the generated data.
\paragraph{Ensuring Robustness} To ensure robustness, the text generation step is performed for three iterations and the results are averaged. Additionally, we perform statistical analysis to check overall whether text generation-based methods are suitable for improving the performance of classifiers or they tend to add more noise versus using no augmentation approaches.

\subsection{Text Classification}\label{classstrategy}
In this final step, we use the augmented training data to train a fastText classifier~\cite{joulin2017bag} coupled with domain-trained fastText word embeddings. 
The reason to use a simple model such as fastText is its efficiency and that transformer-based models tend to not perform well with limited data in document classification and in general tasks that do not require a fine granularity \cite{joshi2020devil}. Indeed, fastText has been shown to perform equally or better with limited labeled data in document classification, compared to more sophisticated models such as BERT \cite{aleks2020edwards}. 

\begin{figure}[t!]
		    \begin{center}
		     \includegraphics[scale = 0.065]{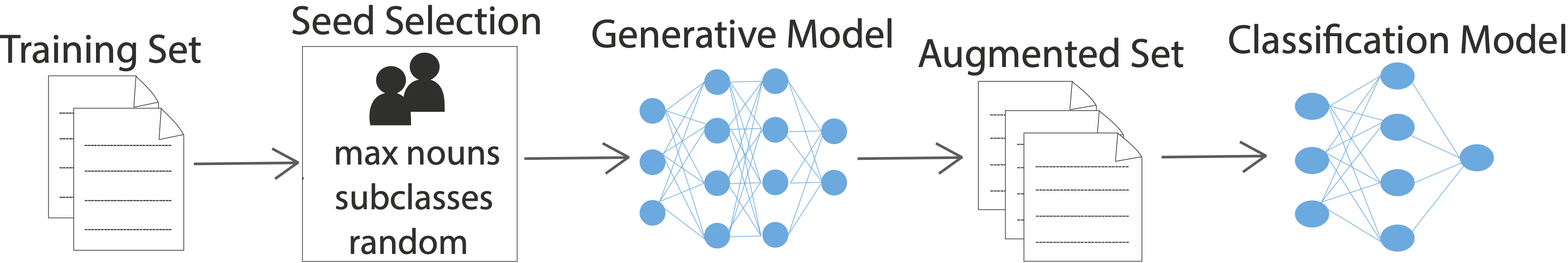}
            \caption{Overview of the methodology}\label{fig:methodology}
		    \end{center}
    \end{figure}

\section{Experimental Setting}\label{sec:settings}
In the following we describe our few-shot text classification experimental setting.\footnote{Code and data are available.}

    \subsection{Safeguarding Domain}\label{ssec:data}
    For our experiments, we selected the \textit{Safeguarding reports} dataset \cite{edwards2020predicting}. The purpose of the safeguarding reports is to identify and describe related events that precede a serious safeguarding incident and to reflect on agencies' roles. As a special trait of this dataset, the reports contain domain-specific terminology which makes them hard to analyse with existing text analysis tools \cite{edwards2019knowledge}.  Further, safeguarding is a multi-disciplinary domain involving terminology and issues from various other disciplines such as criminology, healthcare, and law. Thus, approaches conducted for the safeguarding documents should be applicable for wider range of domains.  Additionally, 
    we perform comparison for two additional datasets which do not require domain expert for annotation. These are: \textit{20 Newsgroups} \cite{lang1995newsweeder} and \textit{Toxic comments} \cite{hosseini2017deceiving} (more information is given in the Appendix). However, we conducted the human-in-the-loop, i.e.,  expert-guided seed selection strategy only for the safeguarding domain where the class framework is created by subject-matter experts. While the manual annotation of the documents is performed on passage level \footnote{Passages in the safeguarding reports are a list of a few sentences which could be viewed as short paragraphs. The labels for the classification remain unchanged.}, we include experiments on sentence level  in order to evaluate performance of text generation methods for generating both short and long sequences.
    We perform prediction for the top classes of the dataset. However, as mentioned in Section~\ref{sec:methodology}, we use the sub-classes to select seed instances. For providing clarity and transparency into the sample generation process, we convert the multi-label classification task of the \textit{Safeguarding} and \textit{Toxic comments} dataset to multi-class problem, removing the few instances that were labeled with more than one class in the original dataset. Focusing on samples with a single label can further help generate stronger class representatives and thus can help both multi-class and multi-label classification. The main features and statistics for the datasets are summarized in Table \ref{safegaurding}.
    


    \begin{table}[!htbp]
\centering
	\resizebox{\linewidth}{!}{
		\begin{tabular}{|l|l|l||c|c|c|c|}\hline
		  \textbf{Dataset}&\textbf{Domain}&\textbf{Task}&\textbf{Class}&\textbf{Subclass}&\textbf{Avg len}&\textbf{\# Test}\\\hline
		  Safeguarding (passages)&Social reports&Theme detection&5&34&45&284\\\hline
		  Safeguarding (sentences)&Social reports&Theme detection&5&34&18&284\\\hline
		   20 Newsgroups&Newsgroups&285&6&20&285&6,728\\\hline
		  Toxic comments&Wikipedia&46&2&5&46&63,978\\\hline
		\end{tabular}
		}
		\caption{Overview of the datasets used for text classification: Average number of tokens per instance (Av len), number of classes (Class), number of subclasses (Subc) and number of test instances (Test)}\label{safegaurding}
\end{table}

\paragraph{Filtering training data.}We focus on few-shot scenarios where the dataset is balanced. We start experiments with 5 and 10 instances per label, extracted randomly from the original data (`base' instances), with at least one instance per subclass. Then, we add 5, 10, and 20 artificially generated instances to the `base' instances (`add' instances) in order to evaluate the effect of methods over different sized training data (consisting of both original and artificially generated samples).

\paragraph{Domain data.}In addition to the datasets with a limited amount of labels, we also leverage domain-specific corpora (in the form of the original training sets for each dataset, without making use of the labels) with two purposes: (1) analyzing the effect on GPT-2 fine-tuned on more data for generating new instances, and (2) recreating a usual scenario in practice, which is having a relatively large unlabeled corpus but a small number of annotations. The corresponding domain corpus were also used by fastText \cite{bojanowski2017enriching} to learn domain-specific embeddings.

\subsection{Text Generation}\label{ssec:training}
    As mentioned in Section~\ref{sec:methodology}, we use the GPT-2 language model \cite{radford2019language} for generating additional training instances. We fine-tuned the GPT-2 model using the GPT-2 Hugging Face default transformers implementation \cite{Wolf2019HuggingFacesTS}. In addition to the pre-trained general-domain model, we fine-tune GPT-2 in each training set as well as per label using causal language model technique where the model predicts the next token in a sequence. We fine-tune the model for 4 epochs and learning rate 5e-5. For generating additional training sequences we use the sampling method of \newcite{holtzman2019curious}.
    
    \subsection{Classification}\label{ssec:classif}
    As mentioned in Section \ref{classstrategy}, we use fastText\footnote{We provide classification results based on fastText trained on the entire non-augmented training sets in the appendix.} as our text classifier \cite[FT]{joulin2017bag} where we use 'softmax', 2 grams, and domain-trained word embeddings. In order to learn domain-specific word embedding models we used the corresponding training sets for each dataset by using fastText's skipgram model \cite{bojanowski2017enriching}. We use fastText word embeddings rather than other word embedding models as they tend to deal with OOV words better than Glove and word2vec approaches. Also, fastText embeddings are the default using the fastText classifier. We report results based on the standard micro- and macro- averaged F1 \cite{yang1999evaluation}.
    

\subsection{Data Augmentation Baselines}\label{ssec:baselines}
For our baselines, we employ synonym, word embedding and language model based strategies for word replacement, and back-translation for sentence replacement (see Section \ref{sec:augmentation} in the Appendix for more details on DA techniques).
As implementations, we rely on {\it TextAttack} \cite{morris2020textattack} for the synonym and word embedding approaches, and {\it nlpaug} \cite{ma2019nlpaug} for the language model and back-translation.
We follow the default configurations for both libraries, where WordNet \cite{miller1998wordnet} is used as a thesaurus for synonym replacement,
BERT \cite{devlin2019bert} ({\it bert-uncased-large}) as the language model, and Transformer NMT models \cite{vaswani2017attention} trained over WMT19 English/Germany corpus for back-translation.

\subsection{Human-in-the-loop Approach}\label{ssec:case}

\begin{figure}[t!]
		    \begin{center}
		     \includegraphics[scale = 0.065]{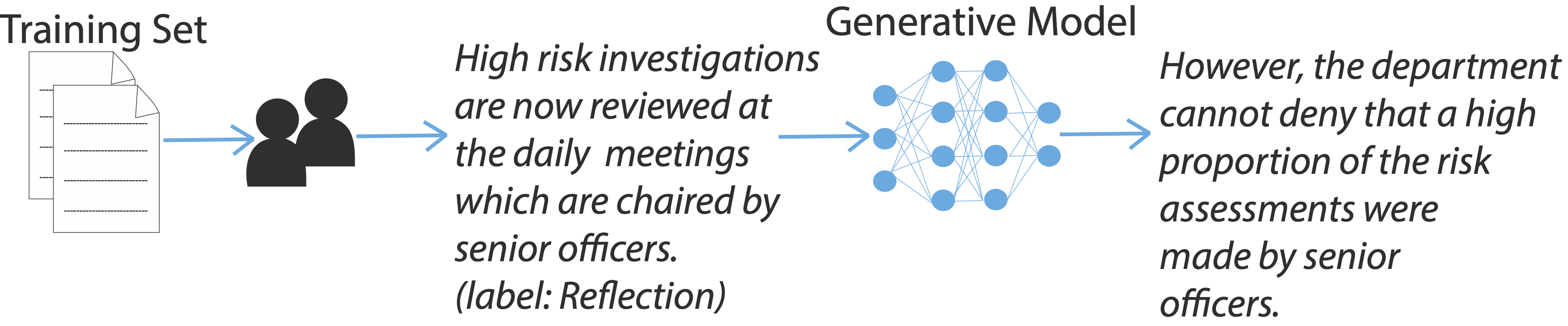}
            \caption{Example of expert-guided seed selection}\label{fig:ex}
		    \end{center}
    \end{figure}

    For the purpose of the experiments, we randomly selected two samples from the original data, one consisting of sentences (`sentence sample') and another one consisting of passages (`passages sample'). Each sample contained 20 instances per label or 100 instances in total. The `sentence sample' and the `passage sample' were distributed among two experts. Participants were asked for each sentence/passage to choose whether it is a \emph{good} or \emph{bad} representative of the class, or to indicate whether they are unsure. We use only a sample of the original data and involve two experts in order to evaluate whether expert-guided seed selection strategy work in a real case scenario in which the selection process is time- and cost- consuming for larger datasets. The experts followed standard procedures in thematic analysis for completing the task, similar to those used for annotating the safeguarding reports \cite{robinson2019making}. Specifically, participants arrived to the final selection of the good theme representative samples through discussion. The participants are practitioners in the safeguarding domain working for Welsh Government, performing qualitative analysis for safeguarding documents. The results from the experiments (see Table~\ref{userexperiments}) show that experts selected more than 10 instances per theme for both samples as `good representatives'. To select 10 and 5 seeds from the `good representatives' we use random selection and max-noun selection strategies. An example of the process is given in Figure~\ref{fig:ex}.
    
    
    \begin{table}[!htbp]
	\centering
	\setlength{\tabcolsep}{4.0pt}
	\resizebox{\linewidth}{!}{
		\begin{tabular}{|l|c|c|c|c|c|}\hline
		\multirow{2}{*}{\textbf{Theme}}&\multicolumn{2}{|c|}{\textbf{passages}}&\multicolumn{2}{|c|}{\textbf{sentences}}\\\cline{2-5}
		&\textbf{\#good rep}&\textbf{\#bad rep}&\textbf{\#good rep}&\textbf{\#bad rep}\\\hline\hline
		Contact with Agencies&12&8&13&7\\\hline
		Indicative Behaviour&12&8&15&5\\\hline
		Indicative Circumstances&11&9&13&7\\\hline
		Mental Health Issues&11&9&14&6\\\hline
		Reflections&11&9&11&9\\\hline\hline
		Total&57&43&66&34\\\hline
		\end{tabular}
		}
		\caption{Results from expert study where `\#good rep' refer to the number of good representative seeds that the expert selected while `\#bad rep' refer to the number of samples that the expert deemed not good representatives of the themes }\label{userexperiments}
\end{table}

\section{Results and Analysis}\label{sec:res} 
    The aims of our analysis is (1) to identify the most suitable method for fine-tuning GPT-2 model to ensure generating higher quality training data (see Section~\ref{gptmodels}), and (2) to understand whether and which seed selection strategies are beneficial for improving DA methods, especially for specialised domains which require domain experts to perform manual annotation (see Section~\ref{gendiscuss}). The results for the three datasets are displayed in Table~\ref{generic}.
    \begin{table*}[t!]
        \centering
		   \resizebox{\linewidth}{!}{
		   \begin{tabular}{|c|c|c|c|c|c|c|c|c|c|c|c|}\hline

		   &\multirow{3}{*}{DA type}&\multirow{3}{*}{Tuning type}&\multirow{3}{*}{DA method}&\multicolumn{4}{|c|}{\textbf{Micro-F1}}&\multicolumn{4}{|c|}{\textbf{Macro-F1}}\\\cline{5-12}
	
		   &&&&\multicolumn{2}{|c|}{5base}&\multicolumn{2}{|c|}{10base}&\multicolumn{2}{|c|}{5base}&\multicolumn{2}{|c|}{10base}\\\cline{5-12}
		   &&&&+5add&+10add&+10add&+20add&+5add&+10add&+10add&+20add\\\hline
            \multirow{9}{*}{20 Newsgroups}&None&-&-&\multicolumn{2}{|c|}{.509}&\multicolumn{2}{|c|}{.578	}&\multicolumn{2}{|c|}{.481}&\multicolumn{2}{|c|}{.567}\\\cline{2-12}
            &\multirow{4}{*}{TG (GPT2)}&gen&random&.539&.536&.572&.555&.519&.519&.564&.548\\\cline{3-12}
		    &&dom&random&.526&.502&.548&.539&.511&.485&.534&.526\\\cline{3-12}
		    &&\multirow{3}{*}{label}&random&\textbf{.609{*}}&\textbf{.602{*}}&\textbf{.627{*}}&\textbf{.637{*}}&\textbf{.591{*}}&\textbf{.587{*}}&.615&.627\\\cline{4-12}
		    &&&nouns&.569&.549&.599&.576&.552&.533&.583&.562\\\cline{4-12}
		    &&&subclass&.563&.585&.624&.632&.549&.571&\textbf{.620{*}}&\textbf{.628{*}}\\\cline{2-12}
		    &\multirow{3}{*}{WR}&-&BERT&.519&.516&.567&.571&.511&.505&.554&.556\\\cline{3-12}
		    &&-&embeddings&.556&.540&.556&.552&.534&.516&.544&.539\\\cline{3-12}
		    &&-&synonyms&.517&.508&.554&.549&.502&.493&.542&.537\\\cline{2-12}
		    &SR&-&translation&.529&.525&.559&.563&.515&.509&.549&.552\\\cline{2-12}
		    \hline \hline\multicolumn{4}{|c|}{\textit{Original data (upperbound)}}&.601&.641&.648&.654&.589&.624&.633&.639\\\hline\hline
		    \multirow{9}{*}{Toxic comments}&None&-&-&\multicolumn{2}{|c|}{.423	}&\multicolumn{2}{|c|}{.442	}&\multicolumn{2}{|c|}{.423}&\multicolumn{2}{|c|}{.442}\\\cline{2-12}
           &\multirow{4}{*}{TG (GPT2)}&gen&random&.447&.424&.405&.423&.447&.424&.405&.423\\\cline{3-12}
	    &&dom&random&.401&.417&.369&.343&.401&.417&.369&.343\\\cline{3-12}
		    &&\multirow{3}{*}{label}&random&\textbf{.453{*}}&\textbf{.452{*}}&.453&.442&\textbf{.453{*}}&\textbf{.452{*}}&.453&.442\\\cline{4-12}
		    &&&nouns&.417&.399&\textbf{.502{*}}&\textbf{.461{*}}&.417&.399&\textbf{.502{*}}&\textbf{.461{*}}\\\cline{4-12}
		    &&&subclass&.427&.440&.419&.421&.427&.440&.419&.421\\\cline{2-12}
		    &\multirow{3}{*}{WR}&-&BERT&.447&.443&.426&.422&.447&.443&.426&.422\\\cline{3-12}
		    &&-&embeddings&.441&.441&.432&.432&.441&.441&.432&.432\\\cline{3-12}
		    &&-&synonyms&.423&.411&.433&.429&.423&.411&.433&.429\\\cline{2-12}
		    &SR&-&translation&.446&-&.436&-&.446&-&.436&-\\\cline{2-12}
		    \hline \hline\multicolumn{4}{|c|}{\textit{Original data (upperbound)}}&.442&.435&.448&.463&.442&.435&.448&.463\\\hline\hline
		    
	\multirow{11}{*}{Safeguard (pass)}&None&-&-&\multicolumn{2}{|c|}{.326	}&\multicolumn{2}{|c|}{.326}&\multicolumn{2}{|c|}{.299	}&\multicolumn{2}{|c|}{.300}\\\cline{2-12}
            &\multirow{6}{*}{TG (GPT2)}&gen&random&.298&.305&.382&.358&.254&.264&.335&.330\\\cline{3-12}
		    &&dom&random&.333&.288&.323&.309&.276&.246&.287&.267\\\cline{3-12}
		    &&\multirow{5}{*}{label{*}}&random&.316&.302&.347&.326&.278&.266&.309&.287\\\cline{4-12}
		    &&&nouns&.375&.337&.375&.379&.329&.281&.338&.351\\\cline{4-12}
		    &&&subclass&.379&.330&.368&.368&.321&.286&.335&.345\\\cline{4-12}
		    &&&expert-random&\textbf{.404{*}}&.386&.393&\textbf{.407{*}}&\textbf{.358{*}}&.349&.342&.352\\\cline{4-12}
		    &&&expert-nouns&.389&\textbf{.435{*}}&\textbf{.410{*}}&\textbf{.407{*}}&.335&\textbf{.382{*}}&\textbf{.351{*}}&\textbf{.366{*}}\\\cline{2-12}
		    &\multirow{3}{*}{WR}&-&BERT&.287&.294&.326&.336&.282&.278&.294&.297\\\cline{3-12}
		    &&-&embeddings&.389&.382&.305&.319&.343&.341&.283&.287\\\cline{3-12}
		    &&-&synonyms&.277&.267&.312&.315&.256&.245&.285&.292\\\hline
		    &SR&-&translation&.333&.336&.298&.312&.294&.301&.273&.286\\\cline{2-12}
		    \hline \hline\multicolumn{4}{|c|}{\textit{Original data (upperbound)}}&.336&.337&.358&.368&.301&.304&.307&.320\\\hline\hline
		     \multirow{11}{*}{Safeguard (sent)}&None&-&-&\multicolumn{2}{|c|}{.242	}&\multicolumn{2}{|c|}{.316}&\multicolumn{2}{|c|}{.193	}&\multicolumn{2}{|c|}{.282}\\\cline{2-12}
            &\multirow{9}{*}{TG (GPT2)}&gen&random&.294&.326&.291&.298&.212&.235&.252&.251\\\cline{3-12}
		    &&dom&random&.298&.326&.291&.302&.214&.236&.252&.250\\\cline{3-12}
		    &&\multirow{5}{*}{label}&random&.295&.326&.291&.302&.213&.235&.251&.252\\\cline{4-12}
		    &&&nouns&.358&.368&.361&.389{*}&.285&.302&.327&.358\\\cline{4-12}
		    &&&subclass&.330&.351&.372&.329&.281&.301&.338&.290\\\cline{4-12}
		    &&&expert-random&\textbf{.337{*}}&\textbf{.375}{*}&\textbf{.361{*}}&\textbf{.414{*}}&\textbf{.298{*}}&\textbf{.336{*}}&\textbf{.340{*}}&\textbf{.379{*}}\\\cline{4-12}
		    &&&expert-nouns&.291&.298&.354&.375&.274&.276&.332&.351\\\cline{2-12}
		    &\multirow{3}{*}{WR}&-&BERT&.249&.284&.319&.315&.245&.274&.278&.274\\\cline{3-12}
		    &&-&embeddings&.242&.280&.316&.319&.226&.259&.276&.283\\\cline{3-12}
		    &&-&synonyms&.256&.266&.319&.326&.241&.256&.281&.288\\\hline
		    &SR&-&translation&.287&.294&.336&.329&.257&.263&.296&.291\\\cline{2-12}
		    \hline \hline\multicolumn{4}{|c|}{\textit{Original data (upperbound)}}&.368&.452&.432&.453&.332&.386&.386&.389\\\hline\hline    
	\end{tabular}

	}\caption{FasText classification results based on Micro-F1 and Macro-F1. Text generation is based on GPT-2, where `gen' refers to the pre-trained general-domain model, `dom' refers to the same model fine-tuned on domain data, and `label', fine-tuned per label. Data is split using 5 or 10  `base' instances per label plus additional 5, 10, or 20 `add' instances, `sent' refers to sentences. The baselines we compare our approaches to are: the word-based replacement (WR) and sentence-based replacement (SR) strategies, `Original data (upperbound)' refers the training data extracted from the original dataset using the same amount of `base' and `additional' instances as for the generative models}.{*} -- Best performing DA methods based on GPT-2 fine-tuned per label lead to statistically significant differences over non-augmented classification (\textit{`None'}) based on t-test results where $p_{value}$ < 0.05.
	 \label{generic}
    \end{table*}

       \begin{figure}[hbt!]
   \begin{center}
        \includegraphics[scale = 0.045]{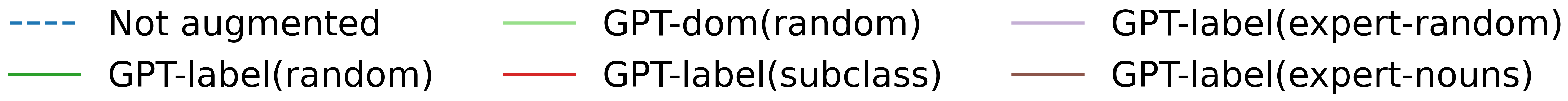}\\\vspace{0.3cm}
        \includegraphics[scale = 0.045]{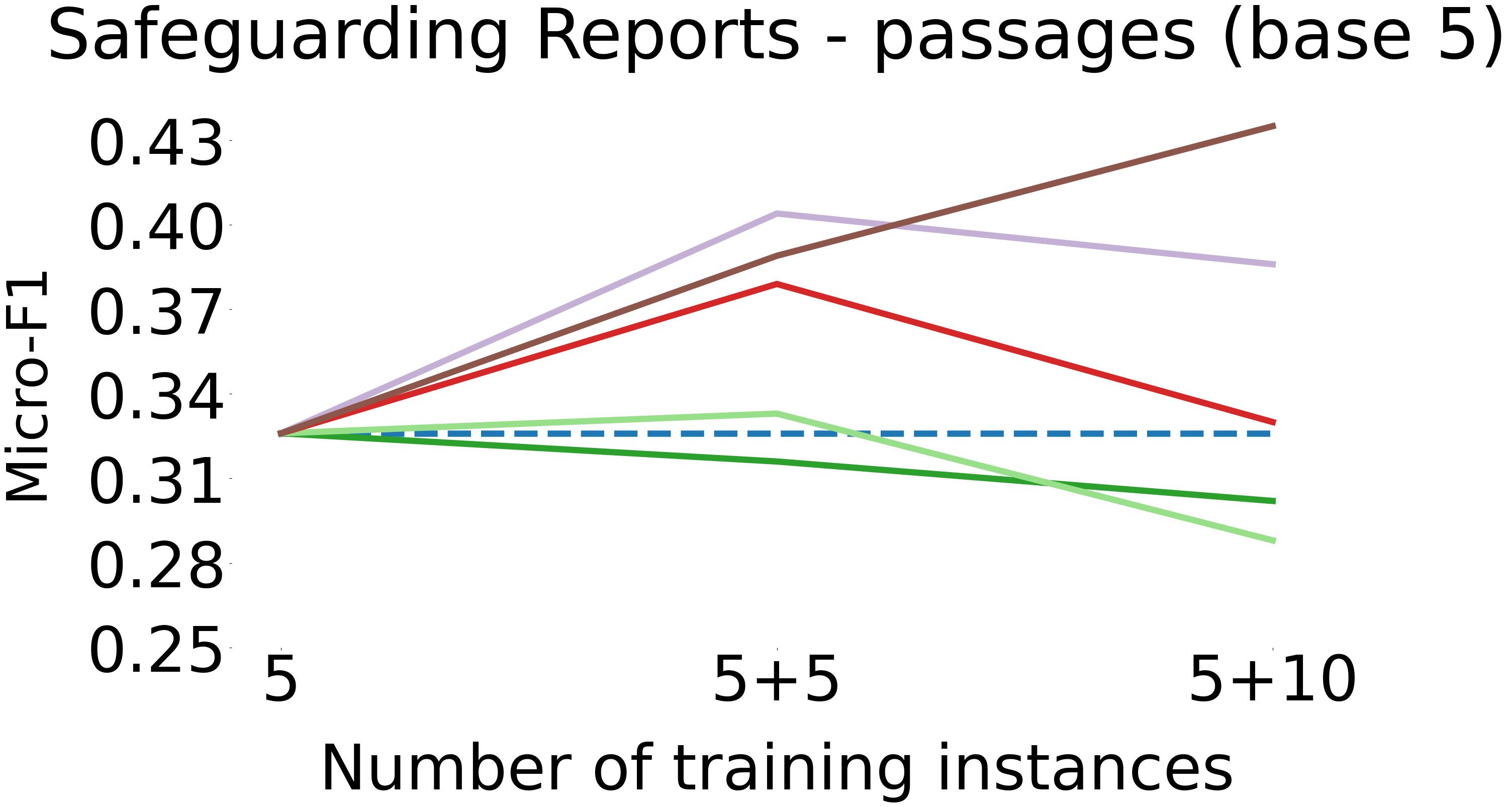}
		\includegraphics[scale = 0.045]{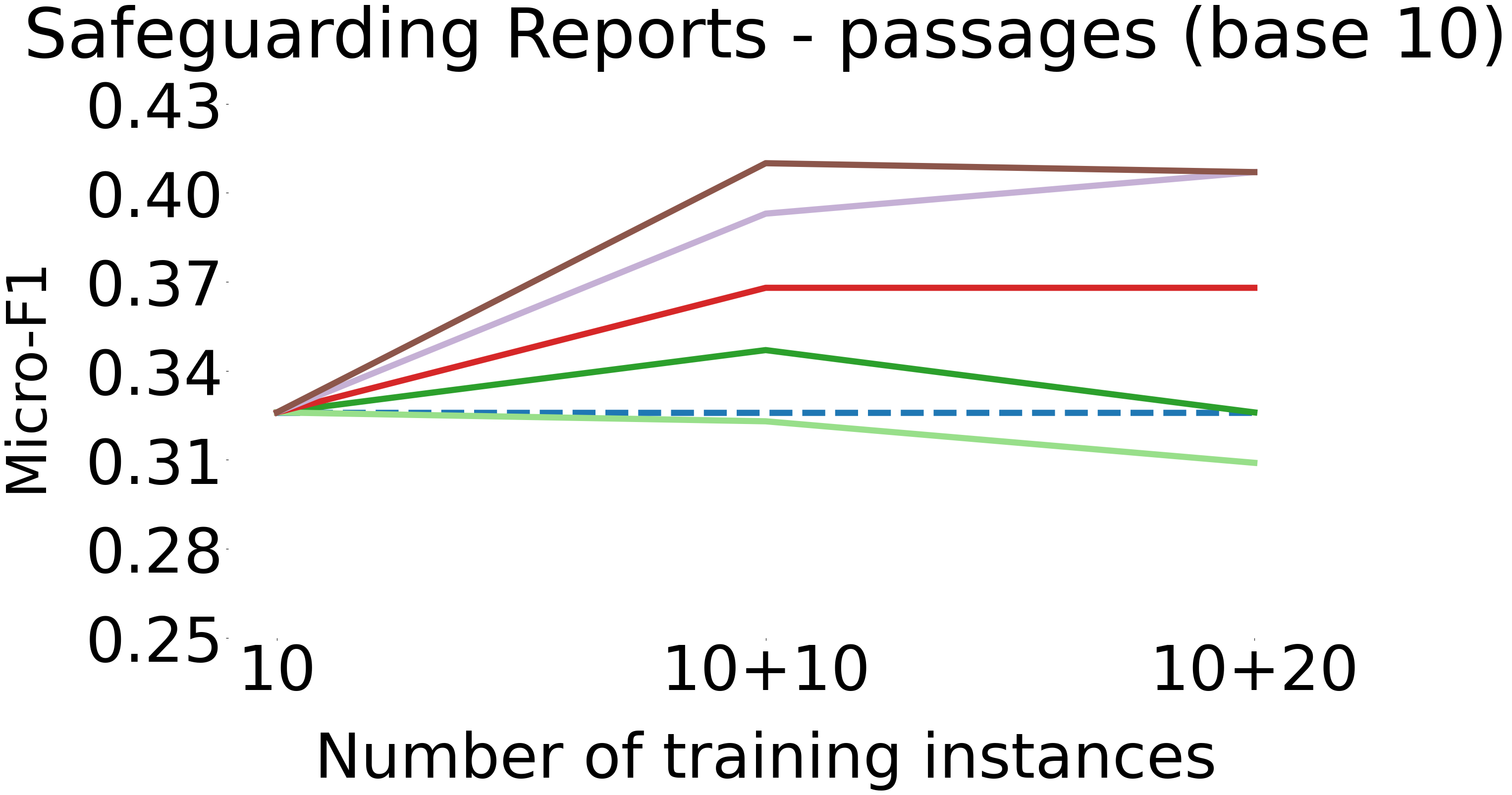}\\\vspace{0.5cm}
		 \includegraphics[scale = 0.045]{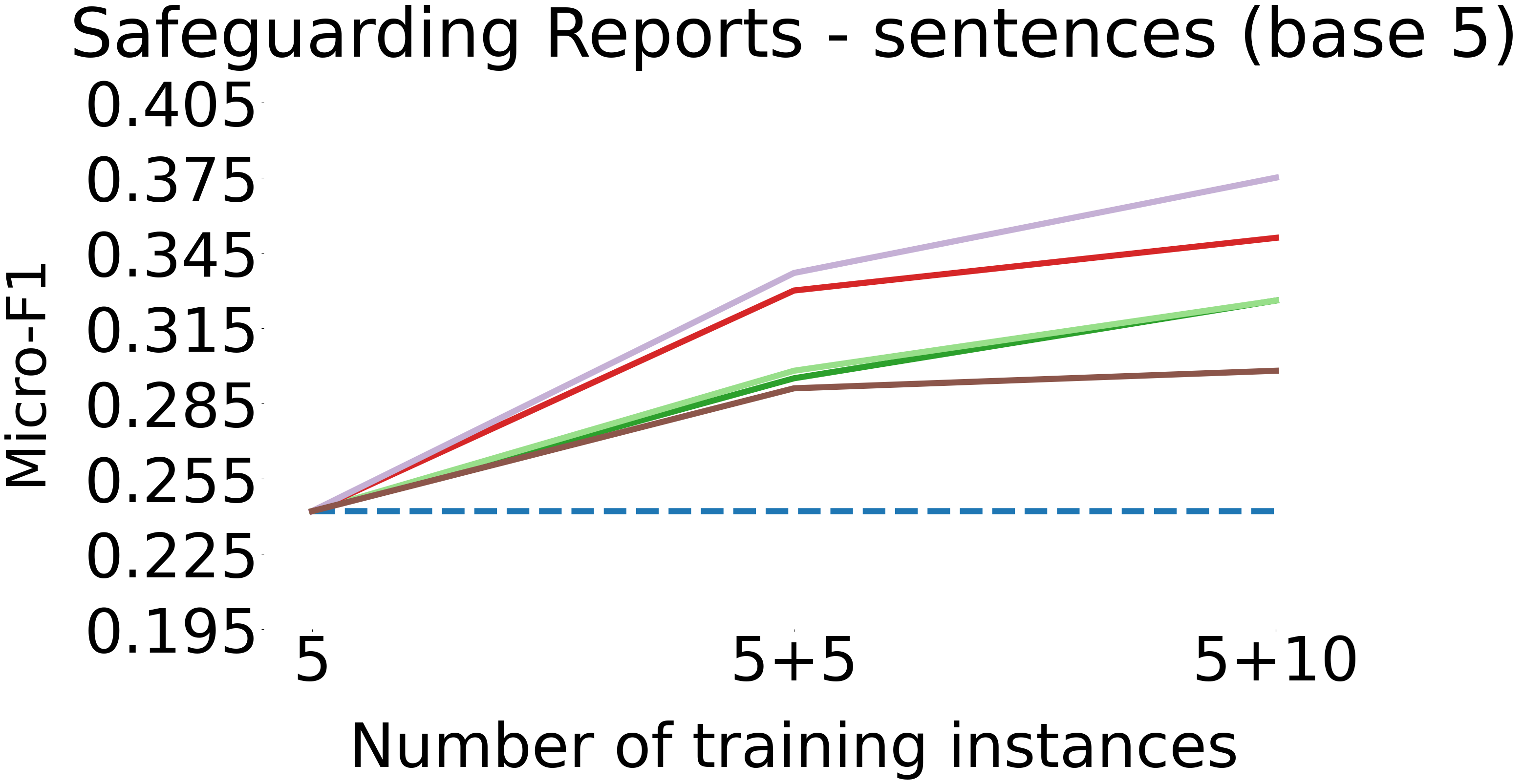}
		 \includegraphics[scale = 0.045]{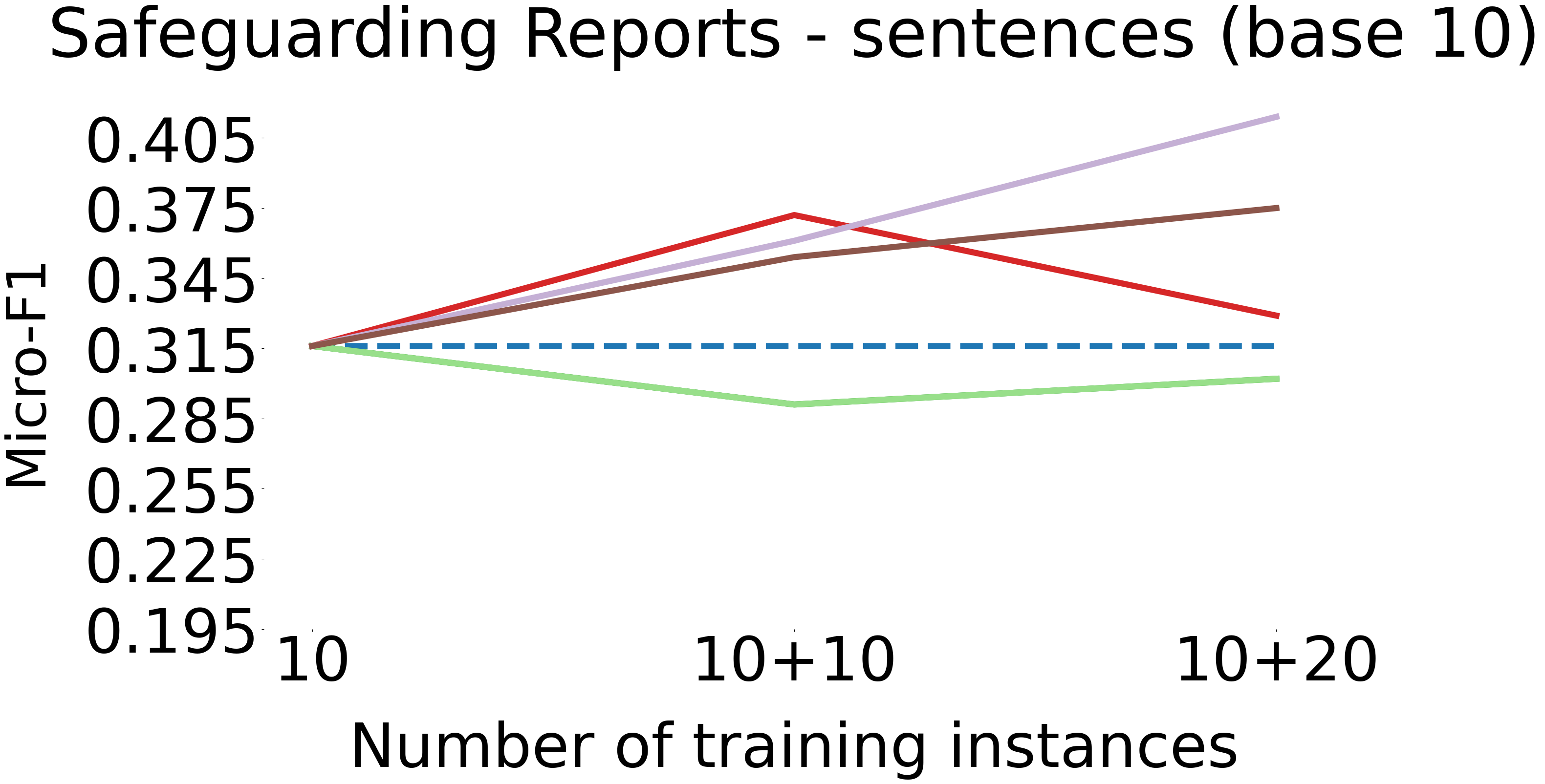}
    \caption{Micro-F1 results with 5 and 10 `base' instances per label for the Safeguarding reports dataset.}\label{microSafeguard}
 \end{center}
\end{figure}
 \begin{figure}[hbt!]
    \begin{center}
    \includegraphics[scale = 0.04]{figures/legendS.png}\\\vspace{0.3cm}
        \includegraphics[scale = 0.04]{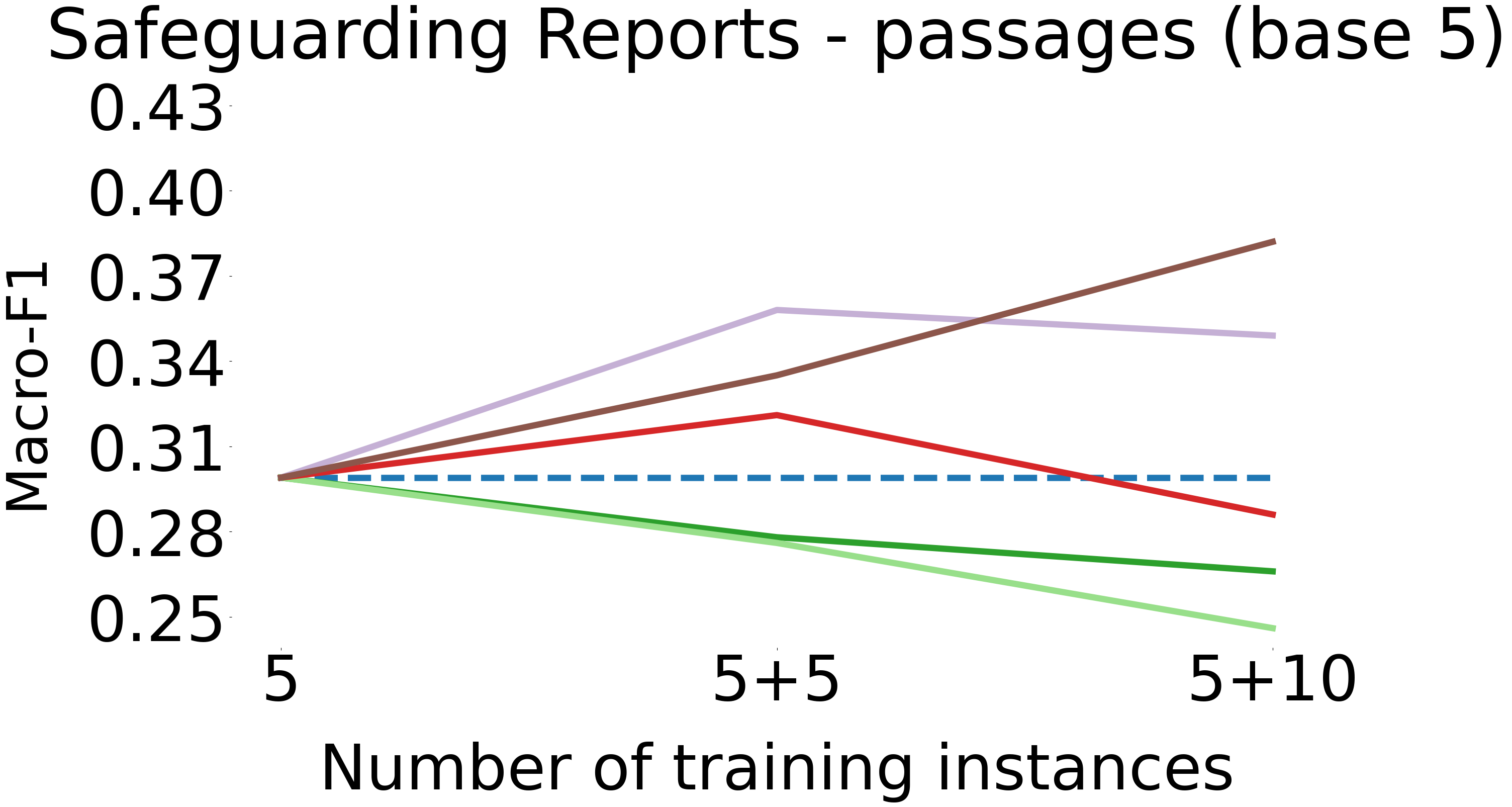}
		 \includegraphics[scale = 0.04]{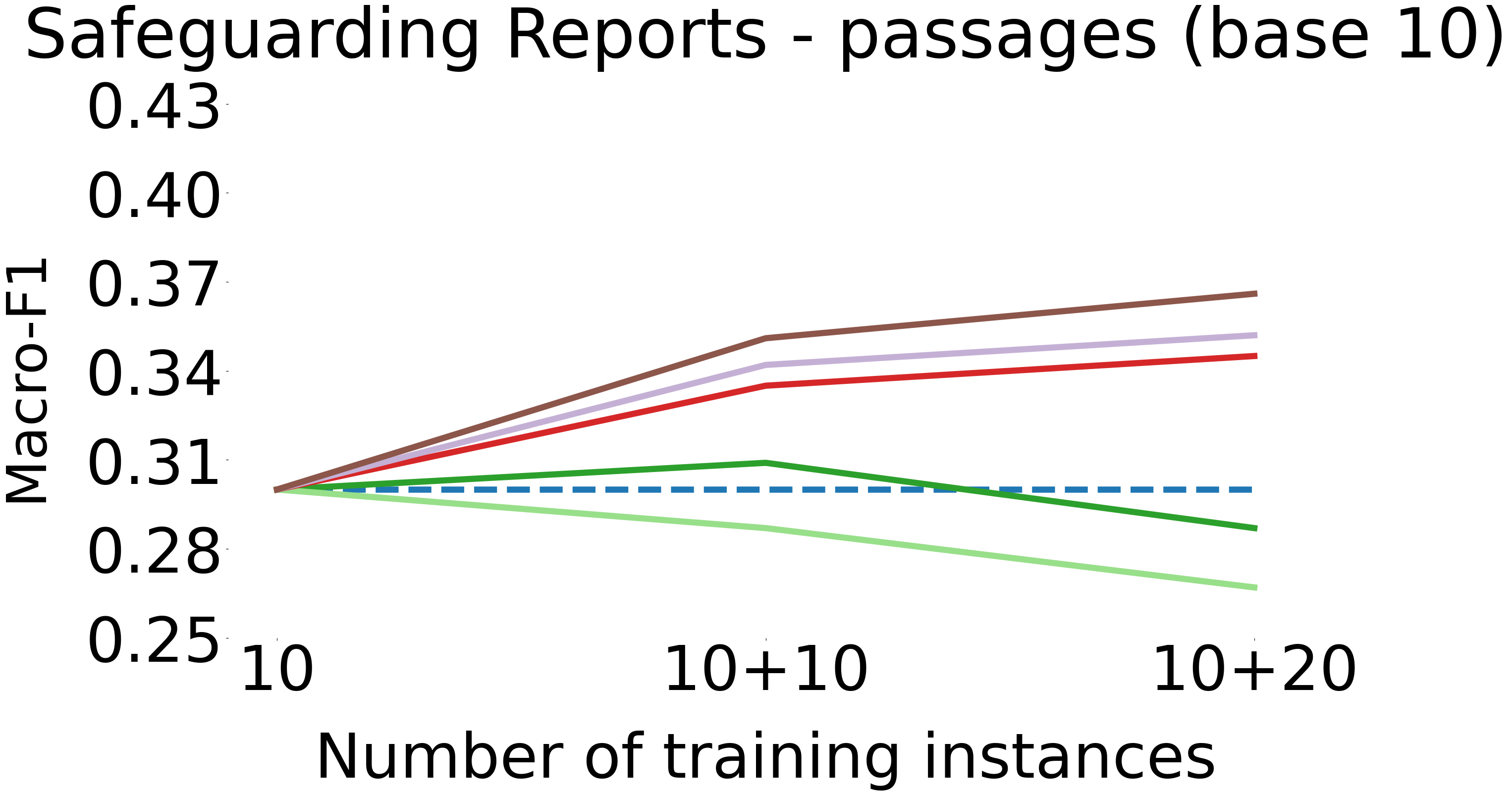}\\
		 \includegraphics[scale = 0.04]{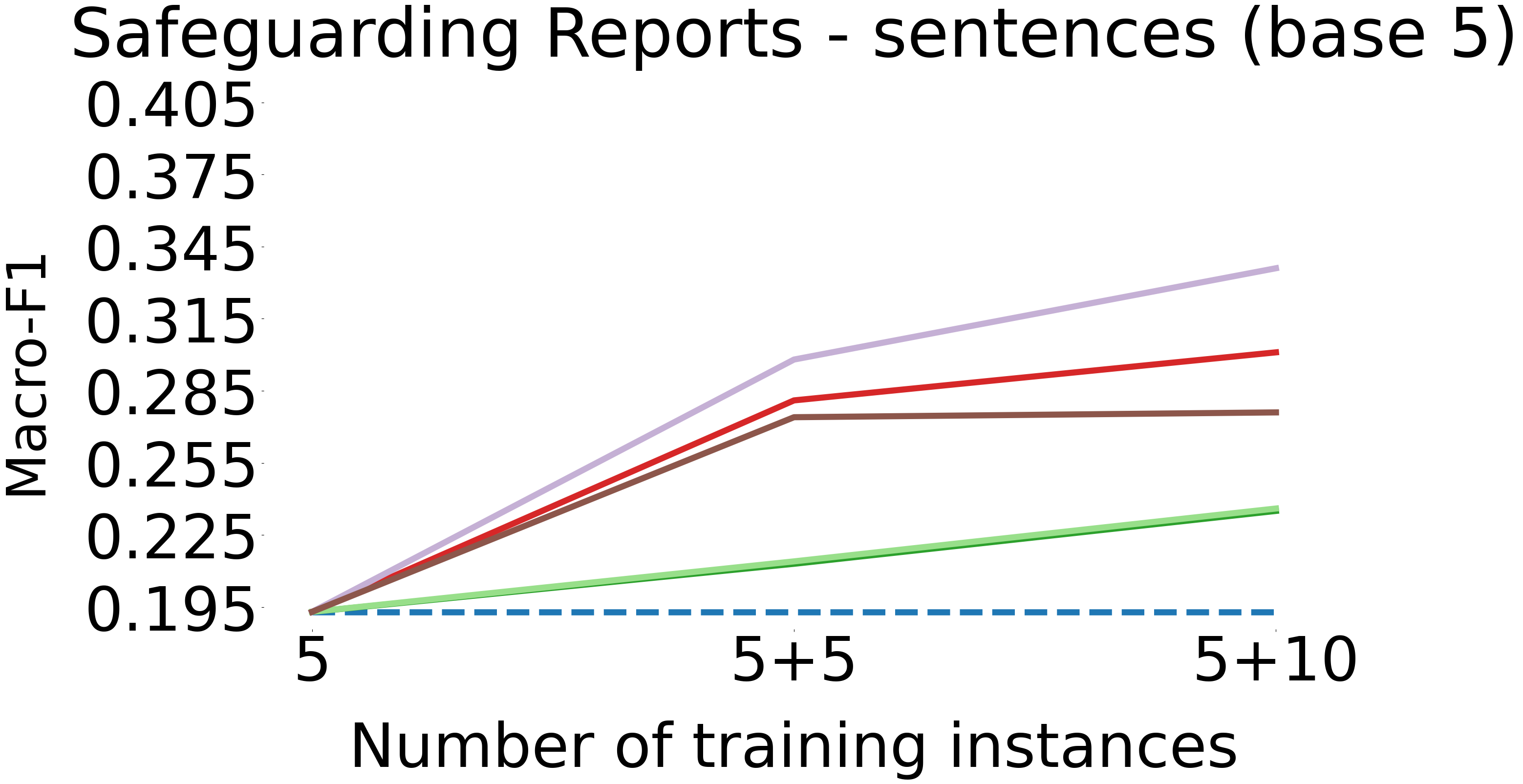}
		 \includegraphics[scale = 0.04]{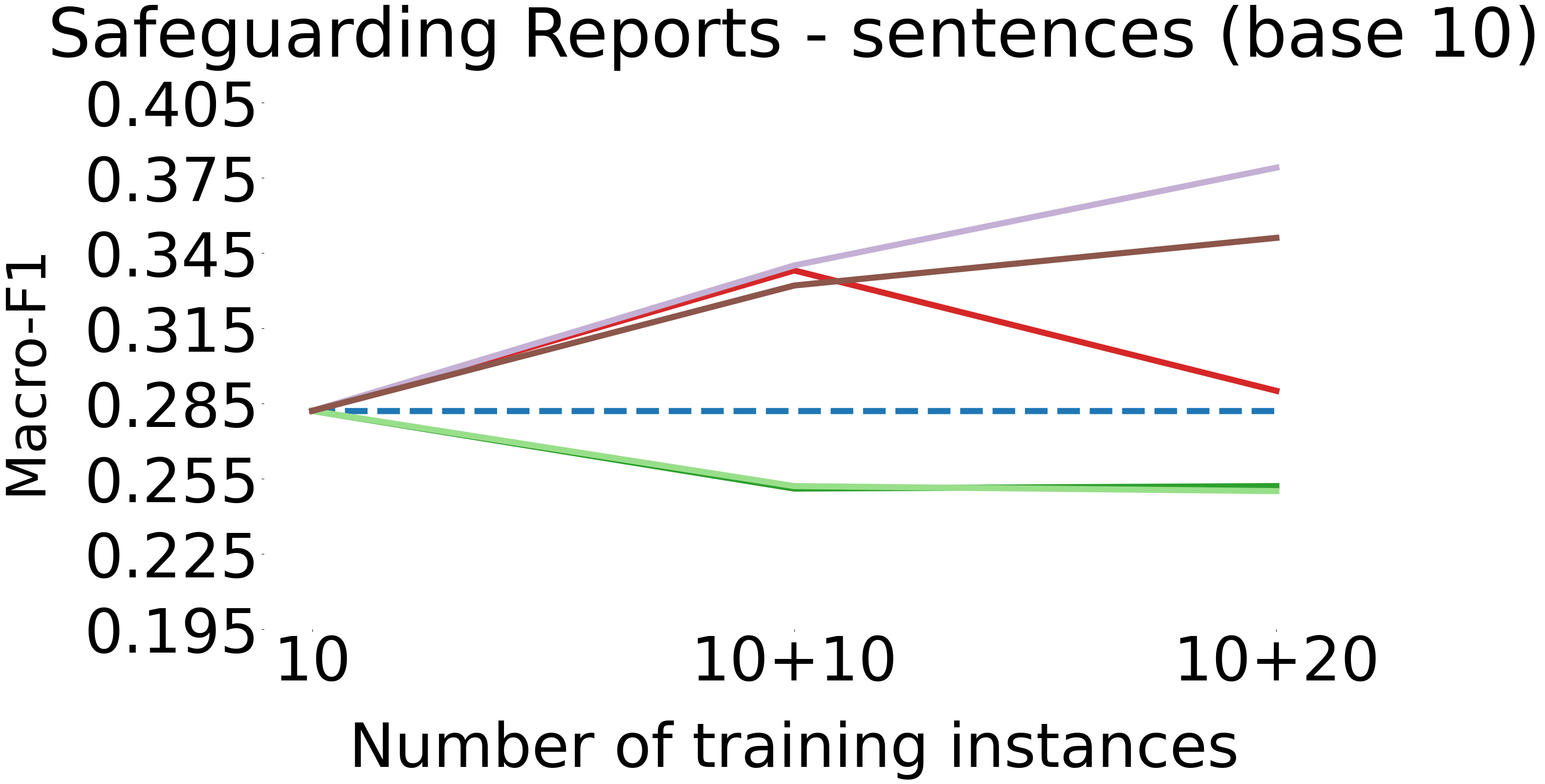}
    \caption{Macro-F1 results with 5 and 10 `base' instances per label for the Safeguarding reports dataset.}\label{allMacroS}
 \end{center}
\end{figure}
 
    \subsection{\textit{Can GPT-based Data Augmentation Help Few-Shot Text Classification?}}\label{gptmodels}

    The results in Table~\ref{generic} indeed confirm the benefits of GPT-based data augmentation. Comparing different methods for fine-tuning GPT-2 models for DA, the classification results show that GPT-2 fine-tuned per label lead to better results, compared to the pre-trained model or GPT-2 fine-tuned on the entire dataset. These results also show that using a fine-tuned GPT-2 model per label does help label-preservation for the generated instances. Surprisingly, the results for the safeguarding reports at the passage level (see Table~\ref{generic}) show that the pre-trained model outperforms the model fine-tuned on the entire dataset for all settings except for `5+5'. This is not the case, however, at the sentence-level where the model fine-tuned on the entire dataset performs very similarly to the model fine-tuned per label. In general, the results clearly suggest that fine-tuning the GPT-2 model on smaller but labelled data works better for classification than fine-tuning it on a larger unlabelled corpus, especially in settings with longer input sequences.  These findings are also supported by the results for the other two datasets,20 Newsgroups and Toxic comments. 
    The main reason for this behaviour can be found in that the fine-tuned model without using label-preservation techniques leads to label-distortions which add noise in the generated dataset. We have given examples of generated instances in the Appendix.
    
   

    \paragraph{Statistical significance tests.} We used t-test~\cite{student1908probable} to measure whether TG-based DA give a significant improvement over the non-augmented classifiers. In particular, we compared the best performing techniques, which are all based on GPT-2 models fine-tuned per label, and the base classifier (`None' in Table~\ref{generic}). We use as a threshold $\alpha$ = 0.05. Results showed that $p_{value} < \alpha$ for every setting. This confirms that fine-tuning GPT-2 model with a small number of labelled instances leads to consistent (and statistically significant) improvements for the safeguarding reports \footnote{These results are also supported by the results for the other two datasets presented in the Appendix}\footnote{We include full results and t-test details in the Appendix.}

    \subsection{Seed Selection Strategies Comparison}\label{gendiscuss}
    Results on comparing seed selection strategies for the specialised domain (i.e., safeguarding reports) (see Figures~\ref{microSafeguard} and \ref{allMacroS}) showed that both seed selection strategies (noun-guided and subclass-guided selection) lead to larger improvements over random selection even for a small number of seed samples. In contrast, experiments on the toxic comments dataset and the 20 newsgroups (see Table~\ref{generic}) showed that random selection is sufficient for improving classification performance over baselines, especially for smaller amount of seeds. This shows that for domains that are similar to the datasets used to train GPT-2 (Newsgroups and Wikipedia) random selection especially for a smaller amount of seeds is sufficient for improving classification performance over baselines. In contrast, applying seed selection techniques to a more specialised domain, such as the safeguarding reports, can be highly beneficial for improving classification.
    
    Finally, the human-in-the-loop approach (see Section~\ref{ssec:case}) revealed that seed selection strategy guided by experts outperform all other seed strategies and baselines for both sentences and passages (see Table~\ref{generic}, Figures~\ref{microSafeguard} and \ref{allMacroS}). This highlights the potential benefits for incorporating expert knowledge into guiding large pre-trained language models in highly specialised domains. This study shows that using active learning techniques in combination with generative models can help increase the efficiency of data augmentation methods and thus be beneficial for few-shot learning. 
 
    
\section{Conclusion}\label{sec:conclusions}
In this paper, we presented and evaluated data augmentation methods using text generation techniques and seed selection strategies for improving the quality of generated artificial sequences and subsequently classifier's performance in few-shot settings. Our results showed that GPT-2 fine-tuned per label, even using only handful of instances, leads to consistent classification improvements, and is shown to outperform competitive baselines and the same GPT-2 model fine-tuned on the entire dataset. This highlights the importance of label preservation techniques in the performance of TG-based DA methods, especially for generating longer sequences (such as passages or full documents).  Seed selection strategies proved to be highly beneficial for the specialised domain analysed in this paper, especially when experts are involved in the selection of class-indicative instances. This shows that combining generative models and active learning techniques, i.e., injecting experts knowledge, can lead to significant improvements in data augmentation methods especially for more specialised domains which require domain experts for the annotation of documents. In future, we plan on expanding the experiments for wider range of specialised domains and compare the performance of bigger generative models such as GPT-3 \cite{brown2020language}, Transformer-XL \cite{dai2019transformer} and CTRL \cite{clive2021control}. Further, we want to investigate what is the optimum amount of artificial training data which can be generated with the described techniques before effecting the classifier's performance negatively.
\newpage
\section*{Limitations}

The main limitation of this research is the lack of further analysis into the performance of text generation models and seed selection strategies when generating higher number of additional training samples. As future work, we plan to investigate the optimal number of generated instances using GPT-based generation as well as experiment with other generative models. Another limitation of the work is that generating artificial training data using GPT-2 requires access to large GPU resources which limits the usability of the approach in real-world scenarios where such resources are unavailable or responses have to be generated in real-time manner. Moreover, the paper presents human-in-the-loop analysis for a single specialised domain (i.e., safeguarding). Safeguarding is a multi-disciplinary domain involving terminology and issues from various other domains such as criminology, medical domain, and legal domain. 
While the results presented in the paper show clear advantage of leveraging expert knowledge into guiding text generation models, we believe that extending the analysis for a wider range of datasets (such as those datasets where we present extended results in the Appendix) can be beneficial. Additionally,
the human-in-the-loop seed selection has been carried by two experts which may cause biases in the process of selecting seeds. However, the participants are practitioners from the safeguarding domain who used standard methodology in thematic analysis for selecting the seeds. These methods do not require inter-annotators agreement, instead experts achieve agreement through discussion. Further, analysis have been performed on sentence- and passage- level where both experiments showed clear advantage of the human-in-the-loop approach.
Finally, the paper presents results for a single high-resource language (English). Experiments for other languages (especially low-resource) could show a different tendency in which the expert involved may be even more necessary.


\bibliography{anthology,custom}
\bibliographystyle{acl_natbib}

\appendix

\section*{Appendix}
\label{sec:appendix}

In Section~\ref{sec:augmentation} we present related research on data augmentation strategies. In Section \ref{overallstats} we describe the classification framework for all three datasets. We also present the statistics for the entire datasets and the classification results using the entire training data per dataset, with no augmentation. 
In Section~\ref{sec:comparison}, we present examples of generated samples between the GPT models we used in our analysis.


\section{Data Augmentation: Related Work}\label{sec:augmentation}
The task of data augmentation consists of generating synthetic additional training samples from existing labelled data \cite{anaby2020not}. In the following, we describe standard text augmentation methods which we use as baselines. We also explain recent DA methods based on text generation models.

\paragraph{Word replacement-based (WR).}\label{ssec:wordrepl}
Simple but commonly used DA techniques are based on word-replacement strategies using knowledge bases \cite{wei2019eda} such as WordNet \cite{miller1998wordnet}. Such methods often struggle to preserve the class label and lead to grammatical distortions of the data \cite{kumar2020data,giridhara2019study,anaby2020not}.  Recent DA approaches address the above issues by using language models to provide more contextual knowledge such as CBERT \cite{wu2019conditional} in the word replacement process. However, methods that make only local changes to given instances produce sentences with a structure similar to the original ones and thus lead to low variability of training instances in the corpus \cite{anaby2020not}. 

\paragraph{Sentence replacement-based (SR).}\label{ssec:sentgen}
Common sentence replacement-based methods are based on back-translation strategies where a given sentence is translated to a language and then back to the original language in order to change the syntax but not the meaning of the sentence \cite{sennrich-etal-2016-improving,fadaee-etal-2017-data}. 

\paragraph{Text Generation (TG).}\label{ssec:textgen}
Recent language models such as GPT-2 \cite{radford2019language} can address the issues associated with the previous strategies by generating completely new instances from given seed samples. 
GPT-2 was trained with a causal language modeling (CLM) objective which makes it suitable for predicting the next token in a sequence. This model has been used successfully in text generation tasks such as summarising \cite{xiao2020modeling,kieuvongngam2020automatic,alambo2020topic} and question answering \cite{liu2019chinese,baheti2020fluent,klein2019learning}. Previous research on using text generation techniques for DA for text classification focused on the creation of label-preservation techniques for the generated synthetic data samples and comparing different TG techniques~\cite{anaby2020not,wang2019classification,zhang2020data,kumar2020data}. However, these works are limited in scale and solutions for improving quality of generated data.Further, There are two main methods used for label preservation of generated samples. The first approach, using a classifier to re-label artificial sequences, requires either  a large training corpus to ensure high performance of the classifier in first place or the generation of large volume of artificial data to ensure that a substantial amount of these will not be filtered because of a low threshold~\citep{anaby2020not}. The other, more widely accepted approach, is prepending the class labels to text sequences during fine-tuning of the Transformer-based model~\citep{wang2019classification, zhang2020data, kumar2020data}. Such an approach cannot ensure label-preservation for all generated sequences. However, our priority is to allow a fair comparison for seed selection approaches without introducing additional noise. Therefore, we consider a simple technique based on fine-tuning a model per label more suitable for performing our analysis.

\section{Datasets description}\label{overallstats}

The 20 Newsgroups collection is a popular data set for experiments in machine learning. The data is organized into 20 different newsgroups, each corresponding to a different news topic such as computer systems, religion, politics \cite{lang1995newsweeder}. The collection of the Toxic comments dataset is obtained from Wikipedia and it is the result from the collaboration between Google and Jigsaw for creating a machine learning-based system for automatically detecting online insults, harassment, and abusive speech~\cite{hosseini2017deceiving}.
Table~\ref{tab:labels} shows that for the 20 Newsgroups dataset there are 20 subclasses split between 6 overall classes. The Toxic comments consists of two overall classes - `toxic' and `non-toxic' where the `toxic' class is overarching 6 subclasses. The Safeguarding reports consists of 5 overall classes and 34 subclasses.

\begin{table}[!ht] 
    \centering
    \scalebox{0.7}{
    \begin{tabular}{|l|p{2.0cm}|p{4.4cm}|}\hline
       \textbf{Dataset} &\textbf{Label}&\textbf{Sub-labels}\\\hline\hline
        \multirow{2}{*}{Toxic comments}&non-toxic&non-toxic\\\cline{2-3}
        &toxic&mild toxic, severe toxic, obscene,threat, insult,identity hate\\\hline\hline
          \multirow{6}{*}{Newsgroups}& computers&comp.graphics, comp.os.ms-windows.misc, comp.sys.ibm.pc.hardware, comp.sys.mac.hardware, comp.windows.x\\\cline{2-3}
        &recreational activities&rec.autos, rec.motorcycles, rec.sport.baseball, rec.sport.hockey\\\cline{2-3}
        &science&sci.crypt, sci.electronics, sci.med, sci.space\\\cline{2-3}
       & forsale&misc.forsale\\\cline{2-3}
        &politics&talk.politics.misc, talk.politics.guns, talk.politics.mideast\\\cline{2-3}
       & religion&talk.religion.misc, alt.atheism, soc.religion.christian\\\hline\hline
        \multirow{5}{*}{Safeguarding Reports}&Contact with Agencies& Health Practitioners, Contact with Third sector orgs, Educational Institutions, Contact with Social Care, Police Contact, Contact with councils or LAs\\\cline{2-3}
        &Indicative Behaviour&Lying, Offending, Serious Threats to Life, Weapons, Emotional Abuse, Domestic Violence, Substance Misuse, Alcohol Misuse, Harassment, Self Inflicted Harm, Stalking, Controlling Behaviour, Aggression\\\cline{2-3}
        &Indicative Circumstances&Bereavement,NFA, Homelessness or Constantly changing Address, Family Structure, Child Safeguarding, Relationship Breakdown, Debt or Financial Exploitation, Sex Work, Relationship with Children, Quality of Relationship\\\cline{2-3}
        &Mental Health Issues&Children, Victim, Perpetrator, Suicidal Ideation \\\cline{2-3}
        &Reflections&Reports Assessments and Conferences, Failures or Missed Opportunities\\\hline
    \end{tabular}
     }
    \caption{Subclasses for the three datasets}
    \label{tab:labels}
\end{table}
The full description of the original datasets is given in Table~\ref{unmodstats}. Results from performing classification using unmodified datasets (using the full training data) are given in Table~\ref{classifentire}. 

 \begin{table}[!t]
    \centering
	   \setlength{\tabcolsep}{4.0pt}
	       \resizebox{\linewidth}{!}{
		  \begin{tabular}{|l||c|c|}\hline
		  \textbf{Dataset}&\textbf{Micro-F1}&\textbf{Macro-F1}\\\hline\hline
		      20 Newsgroups&0.768&0.759\\\hline
		      Toxic comments&0.908&0.908\\\hline
		      Safeguarding Reports (passages)&0.463&0.404\\\hline
		    Safeguarding Reports (sentences)&0.505&0.477\\\hline
	\end{tabular}
	}
		\caption{FastText classification results for the entire datasets with no augmentation.}\label{classifentire}
\end{table}

\begin{table}[!t]
\centering
	\resizebox{\linewidth}{!}{
		\begin{tabular}{|l||c|c|c|}\hline
		  \textbf{Dataset}&\textbf{Avg tokens}&\textbf{\# Train}&\textbf{\# Test}\\\hline
		  Safeguarding Reports (passages)&45&1,261&284\\\hline
		  Safeguarding Reports (sentences)&18&3,591&284\\\hline
		  20 Newsgroups&285&11,231&6,728\\\hline
		  Toxic comments&46&159,571&63,978\\\hline
		\end{tabular}
		}
		\caption{Description of  unmodified datasets}\label{unmodstats}
\end{table}

\subsection{Statistical significance test}\label{ttest}

To further evaluate the effect the additional data generated with GPT-2 have over the classifier's performance, we performed a statistical test, t-test~\cite{student1908probable}, used to compare the means of two groups.
It is used to determine if there is a significant difference between the means of two groups, which may be related in certain features. It is often used to determine whether a process or treatment actually has an effect on the population of interest, or whether two groups are different from one another. 

We use t-test to measure whether the addition of GPT-2 generated training data does actually lead to improvements compared to non-augmented classifier. We specifically perform t-test between best performing seed selection strategy, highlighted in bold  and  `None' row in Tables 3 and 4).  Our $H_0$ is: \textit{Generated data does not lead to overall improvements in classifier performance} and $H_a$: \textit{Generated data does lead to overall improvements in classifier performance}. We use as a threshold $\alpha$ = 0.05. Results in Table~\ref{sigtest} showed that $p_{value} < \alpha$ for every dataset. This confirms that augmenting approaches using seed selection strategies do lead to improvements in classifier's performance versus non-augmented classifiers.

\begin{table}[!ht]
    \centering
	   \setlength{\tabcolsep}{1.0pt}
	       \resizebox{\linewidth}{!}{
		  \begin{tabular}{|l||c|c|c|}\hline
		  \textbf{Dataset}&\textbf{$p_{micro}$}&\textbf{$p_{macro}$}&\textbf{$\alpha$}\\\hline\hline
		      20 Newsgroups&0.01&0.02&0.05\\\hline
		      Toxic comments&0.03&0.03&0.05\\\hline
		      Safeguarding Reports (passages)&0.0001&0.0001&0.05\\\hline
		    Safeguarding Reports (sentences)&0.006&0.016&0.05\\\hline
	\end{tabular}
	}
		\caption{T-test results - compare classification performance with no additional data and results with additional data where performance is the highest}\label{sigtest}
\end{table}

\section{Analysis GPT-2 models: Comparison between generated samples}\label{sec:comparison}

In Table~\ref{examples} we provide examples of generated instances per GPT model. Results showed that the fine-tuned model leads to miss-classifications for the 20 Newsgroup dataset and the Toxic comments dataset. 
    \begin{table*}[!t]
	    \centering
	
	    \resizebox{\linewidth}{!}{

		\begin{tabular}{|p{2.5cm}|p{2.5cm}|p{3.5cm}||p{3.5cm}|p{3.5cm}|p{3.5cm}|}\hline
		  \textbf{Dataset}&\textbf{Label}&\textbf{Original Text}&\textbf{GPT(gen) generation}&\textbf{GPT(dom) generation}&\textbf{GPT(label) generation}\\\hline\hline
		  20 Newsgroups&Computers&There are 3 alternatives: You can try to write a small C program something like: int my\_int\_var;my\_int\_var = system(command);&4. It's a bit confusing and requires very little programming knowledge.5. It uses only your current CPU and some extra memory.&I am now posting the following FAQs to a group of individuals whose answers are from people who are now either atheists, agnostics, etc.&It appears that I am having a problem in trying to use XRemoteAddInput(xmodmap) in windows.\\\hline
		  Toxic Comments&non-toxic& I think the recent editing by all has been brilliant. If I see anything that could be improved I'll speak up here. I would like to see a Christian Theosophy section also. &Thank you for your help, H, for your work on my blog.You are welcome to submit comments in this section, but please note I am not allowed to comment on other forums because they may interfere with my work.&No need to be an encyclopedic asshole. You don't see me deleting things because of something that was on my watchlist. I just wanted you to know that my name is now on my watchlist, and I will try to improve it. &I hope to see the article finished as soon as possible. But I would like to thank you for your contributions. \\\hline
		  Safeguarding reports&Indicative Circumstances& Medical records show that she struggled for many years with depression, for which she was treated with anti-depressants, also receiving referrals for counselling. She also had issues with her physical health.&Her lawyer, Richard Gage, has asked the court to intervene if Mr. McDevitt was not criminally liable in his decision.&She attended at her GP with a range of depressive symptoms and her medication was improved. It is not known how she gained her weight but it is said that her mental health declined and she started following a diet with vitamin C, which was associated with being better at controlling her mood.&In the early 1970s she moved from Los Angeles to New York. During this time she suffered from depression and other problems, but there was little support in New York.She moved into a small apartment and was living with friends, but had to move into a house next to her apartment to support her mental health.\\\hline

		\end{tabular}
		}
	\caption{Examples of generated samples using GPT-2 models, where Safeguarding Reports examples are non-verbatim due to data sensitivity.}\label{examples}
    \end{table*}
    
\end{document}